\documentclass[letterpaper, 10 pt, conference]{ieeeconf}
\IEEEoverridecommandlockouts
\usepackage{cite}
\usepackage{amsmath,amssymb,amsfonts}
\usepackage{graphicx}
\usepackage{textcomp}
\usepackage{xcolor}
\usepackage{graphicx}
\usepackage{tikz}
\usepackage{booktabs} 
\usepackage{algorithm}
\usepackage{hyperref}
\usepackage{verbatimbox}
\usepackage[noend]{algorithmic}

\newcommand{\ra}[1]{\renewcommand{\arraystretch}{#1}}
\newtheorem{definition}{Definition}[section]
\usepackage[caption=false, font=footnotesize]{subfig}

\def\BibTeX{{\rm B\kern-.05em{\sc i\kern-.025em b}\kern-.08em
    T\kern-.1667em\lower.7ex\hbox{E}\kern-.125emX}}
\begin{document}
\begin{minipage}{0.7\paperwidth}
\begin{center}
This paper has been accepted for publication in IEEE International Conference on Robotics and Automation.

\vspace{5mm}

\vspace{10mm}
Please cite our work as:

\vspace{5mm}
V. Cavinato, T. Eppenberger, D. Youakim, R. Siegwart and R. Dub\'e. ``Dynamic-Aware Autonomous Exploration in Populated Environments.'' IEEE International Conference on Robotics and Automation (ICRA), 2021.
\end{center}

\vspace{10mm}
bibtex:
\begin{verbnobox}[\small]
@inproceedings{cavinato2021dynamicaware,
  title      =  {Dynamic-Aware Autonomous Exploration in Populated Environments},
  author     =  {Cavinato, Valentina and Eppenberger, Thomas and Youakim, Dina and 
                 Siegwart, Roland  and Dub{\'e}, Renaud},
  booktitle  =  {IEEE International Conference on Robotics and Automation (ICRA)},
  year       =  {2021}
}
\end{verbnobox}
\end{minipage}

\title{\LARGE Dynamic-Aware Autonomous Exploration in Populated Environments
}

\author{Valentina Cavinato\textsuperscript{1, 2}, Thomas Eppenberger\textsuperscript{1}, Dina Youakim\textsuperscript{1}\thanks{$^1$ Sevensense Robotics AG, Zurich, Switzerland}, Roland Siegwart\textsuperscript{2}\thanks{$^2$ Autonomous Systems Lab (ASL), ETH Zurich, Switzerland}, Renaud Dubé\textsuperscript{1}\thanks{Email address:  \href{mailto:cavinatovalentina@gmail.com}{cavinatovalentina@gmail.com}}
}
\maketitle

\begin{abstract}
Autonomous exploration allows mobile robots to navigate in initially unknown territories in order to build complete representations of the environments. In many real-life applications, environments often contain dynamic obstacles which can
compromise the exploration process by temporarily blocking passages, narrow paths, exits or entrances to other areas yet to be explored.
In this work, we formulate a novel exploration strategy capable of explicitly handling dynamic obstacles, thus leading to complete and reliable exploration~outcomes in populated environments.
We introduce the concept of \textit{dynamic~frontiers} to represent unknown regions at the boundaries with dynamic obstacles together with a cost function which allows the robot to make informed decisions about when to revisit such frontiers.
We evaluate the proposed strategy in challenging simulated environments and show that it outperforms a state-of-the-art baseline in these populated scenarios.
\end{abstract}
\section{Introduction} \label{sec:introduction}
Demand for autonomous service robots such as healthcare, warehousing or cleaning robots is in continuous rise~\cite{increasing-demand, healthcare}. These robots often rely on an initial mapping session prior to being deployed for performing their task autonomously. Autonomous exploration is thereof an important robotics capability, relieving operators from mundane tasks such as teaching or remapping of changing environments, thus reducing operational costs and improving effectiveness.
\begin{figure}
    \centering
    \includegraphics[width=0.8\linewidth]{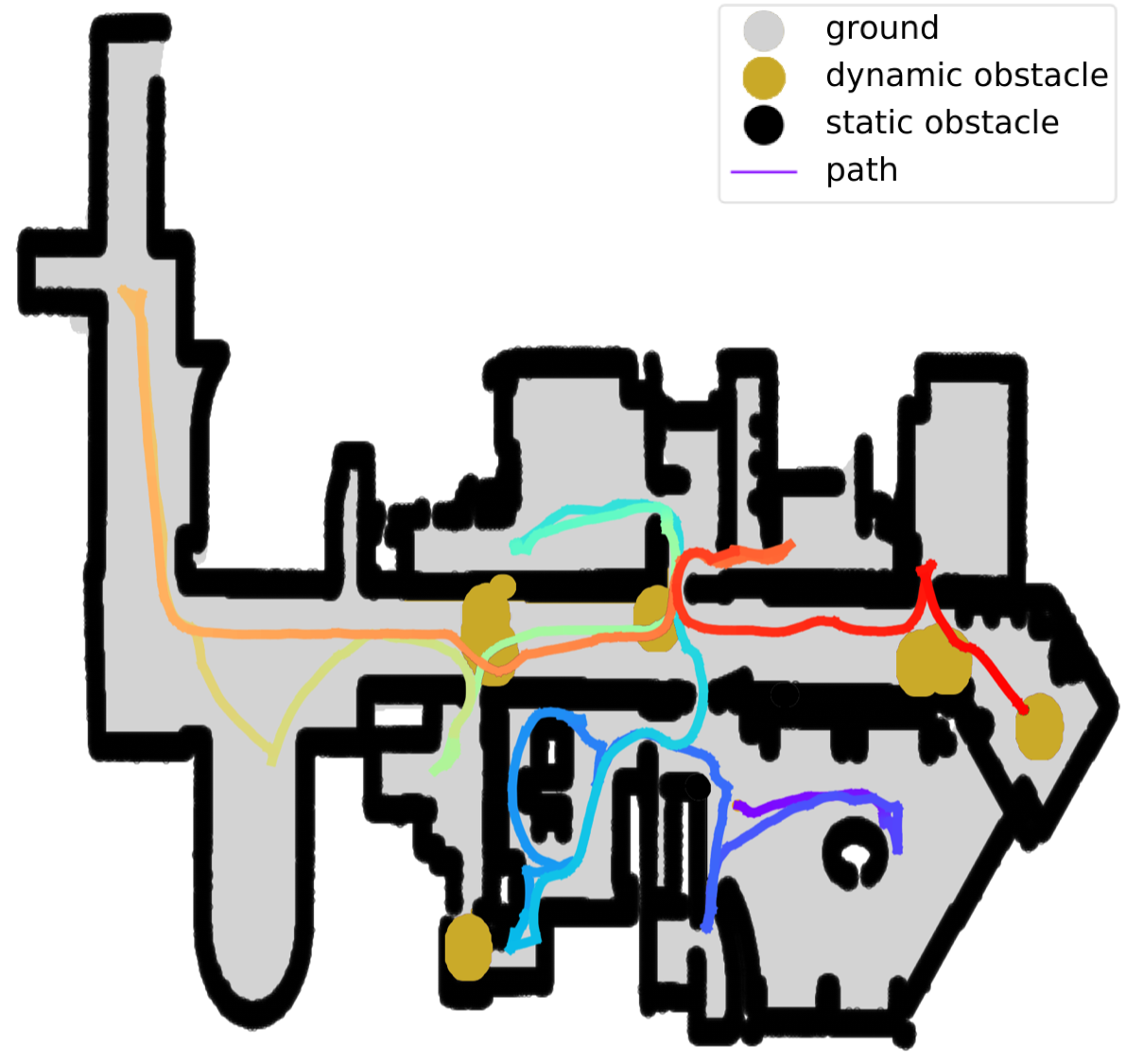}
    \caption{Demonstration of the proposed dynamic-aware exploration strategy in a simulated hospital-like environment containing nine moving people. The final map generated at the successful completion of the exploration is shown with ground cells in grey and static obstacle cells in black. People's positions and dimensions when the robot explored that part of the environment are shown in yellow. The path followed by the robot is also indicated, from its initial position (blue) to its finish point (red) in a rainbow colormap. The robot encountered in its path five passages blocked by people and was able to continue and complete the exploration. }
    \label{fig:teaser}
\end{figure}
Different families of strategies have emerged through the years to tackle this challenge. These differ in the way of identifying the remaining areas to explore~\cite{frontier, next-best-view, ros-ch}, the adopted map representation (occupancy grid~\cite{occupancy-grid}, feature maps~\cite{sonar, feature-based}, pose-graph~\cite{pose-graph} etc.), the use of a single or multiple robots~\cite{frontiers-coord, multirobot, multirobot2} and the objective of the strategy. Some techniques optimize for coverage~\cite{coverage}, other ones try to reduce the localization uncertainty~\cite{reduce-uncertainty}, or balance coverage, accuracy, and speed in an integrated strategy~\cite{information-gai}.  Among the most popular approaches, frontier-based exploration~\cite{frontier} suggests that a robot's best option to increase the knowledge of its surroundings is travelling to the boundaries between known and unexplored regions~(frontiers).
 On the other hand, sampling-based approaches sample the configuration space and select the next robot's pose for example by maximizing the number of map cells that could change occupation status while the robot reaches the designed pose \cite{next-best-view, ros-ch}.
 
Interestingly, we could only identify few works which deal with exploration of populated environments~\cite{surprise, jojo, populated-bayesian, populated}. Hospitals, offices, warehouses or industrial settings in general are all examples of such environments where the coexistence of robots and other moving machines or people may occur. Indeed, it could be impossible for some of these buildings to be completely emptied from moving obstacles for the time needed to a robot to map the entire environment which is particularly unfavorable if changes happen frequently and the procedure has to be repeated often. In these cases, designing an approach capable of dealing with \textit{dynamic obstacles} is fundamental to enable reliable and complete exploration and open a wide range of new applications.
The few existing approaches that we could identify in this direction require interaction with the agents populating the environment~\cite{jojo, populated-bayesian} or make strong assumptions on them, i.e. the agents are capable of effectively moving in crowded spaces~\cite{populated}.

This work presents an efficient exploration strategy which targets populated environments by explicitly handling generic dynamic obstacles. To the best of our knowledge, this is the first work to distinguish between static and \textit{dynamic~frontiers} and leverage this information in the exploration strategy. By integrating the new concept in a cost-based approach we are capable of dealing with challenging scenarios and achieve successful exploration even when dynamic obstacles are moving in the subject environment. A demonstration of the proposed approach is shown in Fig.~\ref{fig:teaser}, and a video presenting an overview of the work and further visualizations is available at \url{https://youtu.be/Mk1xz4AK9MM}.

To summarize, this paper presents the following contributions:
\begin{itemize}
\item A novel dynamic-aware approach for autonomous and efficient exploration of populated environments.
\item A cost function allowing robots to decide on where to travel next: standard frontiers or the newly introduced \textit{dynamic frontier} ones. 
\item An evaluation of the proposed strategy in challenging simulated environments populated by moving actors. 
\end{itemize}
The paper is structured as follows: Section II provides an overview of related work in frontier-based exploration and the few examples of strategies in populated environments. Our approach is then described in Section III and evaluated in Section IV. Finally, Section V draws the conclusions.


\section{Related Work}\label{sec:related-work}
The proposed strategy evolves from the frontier-based exploration first introduced by Yamauchi~\cite{frontier}. Using an occupancy grid that is updated with new sensor measurements, regions at the boundaries between open and unknown spaces are detected. These are called \textit{frontiers}. Suitable travel points are then computed, e.g. by determining the centroid of the frontier or its closest point to the robot, and the best next location is selected as a goal. To detect frontiers, classic computer vision edge-detection methods are used, for example in~\cite{frontier, computer-vision}. Instead, to achieve faster detection, Keidar and Kaminka~\cite{wfd} propose a Wavefront Frontier Detector (WFD) performing a graph search over the known portions of the environment and a Fast Frontier Detector (FFD) that processes only new sensor readings. Among the detected candidates, the next-best-view is then selected based on different criteria. For instance, the original approach~\cite{frontier} selects the closest frontier to the robot, suggesting that new information to be acquired is the same for all detected frontiers. The work in~\cite{information-gain} evaluates the utility of a candidate by estimating the knowledge gain of the query location through probabilistic models. Vutetakis and Xiao~\cite{loop-closure} propose to integrate information gain including loop-closure planning in the selection of next best view candidates. Instead, the strategy proposed in~\cite{scaramuzza} selects as goal the frontier that is in the field of view of the robot and requires the smallest change in velocity, in order to allow consistently higher speeds. The proposed approach is derived from \textit{closest} frontier~based strategies. These have been demonstrated competitive
to other next-best-view strategies and offer a very good trade-off between computation and performance~\cite{comparison, comparison_1}.
In addition, by recomputing the next goal every time new information on the environment is available, we avoid cases in which the robot starts moving towards a previously selected goal which is no longer a region of interest. Starting from the idea in~\cite{wfd}, we only scan the free space until the closest frontier is found.\\
The aforementioned strategies are not evaluated in and do not include nor mention the challenge of a populated environment. To the best of our knowledge there exists only few works in this direction. The work in~\cite{surprise} proposes an exploration carried out by agents motivated by human-like emotions related to exploration such as surprise, curiosity, and hunger. It considers environments populated with entities which can be buildings, objects or other animated agents. However, the effect dynamic objects have on the exploration is not evaluated.
In contrast, an example of human-robot interaction strategy is proposed in~\cite{jojo}, which introduces the Jijo-2 robot. The robot learns meaningful features and a probabilistic map of its environment by talking to the people around and asking for directions. Similar to this, Lidoris et al.~\cite{populated-bayesian} propose a strategy for getting the robot to a destination goal in an unknown environment which includes approaching the detected people to ask for directions. The work introduces a Bayesian framework allowing recursive estimation of the environment's dynamic model and action selection. A more recent work proposes instead a mixed and interactive strategy~\cite{populated}. It exploits human natural heuristics as understanding of a crowded scene and ability to easily walk through a dense environment. The robot has the choice of exploring frontiers or selecting a human to follow while its goal remains exploring a possibly populated environment with minimum distance and time. The strategy has been shown in simulation to be beneficial in some cases.
Contrastingly to the aforementioned works, our approach does not rely on the interaction with moving entities and is not limited to humans as dynamic obstacles as it is agnostic to their nature. The details are provided in the next section.
\section{Approach}
\begin{figure*}[t]
\centering
\includegraphics[width=1.0\linewidth]{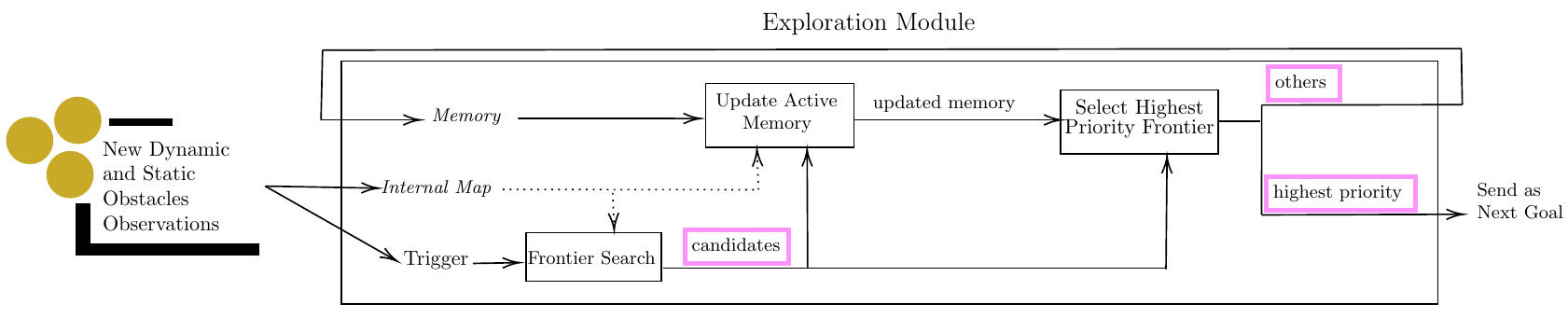}
\caption{Overview of the exploration module and the main components of the cost-based dynamic-aware exploration strategy. When new observations relative to static or dynamic obstacles are available, they are incorporated in the exploration module's internal map and trigger a frontier search. A memory is maintained with the past $dynamic$, $mixed$ and $mixed~simple$ frontiers that were not selected as goals. Candidate frontiers found at the current iteration are used to update the memory together with the information contained in the internal map. Then, the updated memory and the current candidates are scored according to their cost value and the highest priority frontier sent as goal to the navigation module. Other unselected frontiers are added to the memory. A pink square indicates that the output is a list of frontiers.}
\label{fig:overview-module}
\end{figure*}

We propose a dynamic-aware frontier-based exploration strategy with a custom scoring function to select the next goal. As introduced in Section~\ref{sec:related-work}, frontiers represent regions at the boundaries between free and unknown spaces. During the exploration procedure, a frontier is selected as next travel goal at every iteration until none is left and the exploration is considered complete. An overview of the main components of our exploration module is shown in Fig.~\ref{fig:overview-module}.
 The module maintains an internal map by incorporating new knowledge from depth sensors and information on dynamic obstacles provided by an object detection module. This map has the form of a 2D costmap where each cell can be in one of four states: \textit{unknown}, if no information is yet available, \textit{free}, if it has been observed as traversable ground, occupied by a \textit{static obstacle}, or occupied by a \textit{dynamic obstacle}. The map is used to extract frontiers in the goal selection procedure and to update the memory, defined in Section~\ref{subsec:goal-selection-strategy}. 
  Compared to previous works, that followed from~\cite{frontier}, in this paper we extend the classic definition of frontier as presented in the next section.

\subsection{Terminology}

\begin{definition}{\textit{Frontier Cell: }}\label{def:fc-dyn} An unknown cell with either free or dynamic obstacle cells neighbours. We separate two types of frontier cells:
\begin{itemize}
\item \textit{Simple}: if the unknown cell has at least one neighbouring free cell and no static nor dynamic obstacle cell neighbours.
\item \textit{Dynamic}: if the unknown cell has at least one neighbouring dynamic obstacle cell and no neighbouring static obstacle cells.
\end{itemize}
\end{definition}
\begin{definition}{\textit{Frontier:}}\label{def:dyn-f} a set of connected frontier cells. Since two types of frontier cell exist, we define a type for each frontier. With $size_s$ as the number of simple frontier cells in the frontier, $size_d$ the number of dynamic ones,  $sd\_ratio = \frac{size_s}{size_d}$ and the threshold $thresh$:
\begin{equation}\label{eq:dynamic-frontiers}
type=
\begin{cases}
  simple, & \text{if } size_d == 0 \ \\
  mixed \ simple, & \text{if } sd\_ratio \geq thresh \\
  mixed, & \text{if } \frac{1}{thresh} \leq sd\_ratio < thresh \\
  dynamic, & \text{if } sd\_ratio \leq \frac{1}{thresh}
\end{cases}
\end{equation}
Visualizations of the different types are presented in Fig.~\ref{fig:frontiers}.
\end{definition}

\subsection{Goal selection strategy} \label{subsec:goal-selection-strategy}

Algorithm \ref{alg:frontier-search-dyn} shows the proposed approach. At the core, it is a closest frontier strategy: the robot should visit next the frontier at minimum distance as explained in Section \ref{sec:related-work}. However, if the latter is of a type different from $simple$, the frontier search continues until the closest $simple$ frontier is found. Frontiers of type $dynamic$, $mixed$ or $mixed~simple$, may not be good candidates for the next location to visit: the robot should attempt to stay clear of dynamic objects as its interactions with these entities would hardly be predictable and may lead to loss of localization. In addition, the location of dynamic actors is likely to change over time together with the associated frontier's type. Thus, it is likely to be safer to postpone their exploration. The described frontier search is implemented as a breadth-first search (BFS) from the robot's current position (line \ref{line:bfs-dyn}), given the latest available map of the environment.  
All detected frontiers are compared based on their cost (line \ref{line:ascending}, ignoring the $memory$ for now as if it were empty). The cost function reflects the above reasoning such that the cost of a frontier:
\begin{itemize}
    \item Increases with increasing distance from the robot.
    \item Is initially high for frontiers of type $dynamic$, $mixed$ and $mixed~simple$, to encourage the robot to stay away from dynamic obstacles.
    \item Is higher the higher the number of dynamic cells.
    \begin{figure}[H]
\centering
\subfloat[Frontier of type $simple$.]{\includegraphics[width=0.5\linewidth]{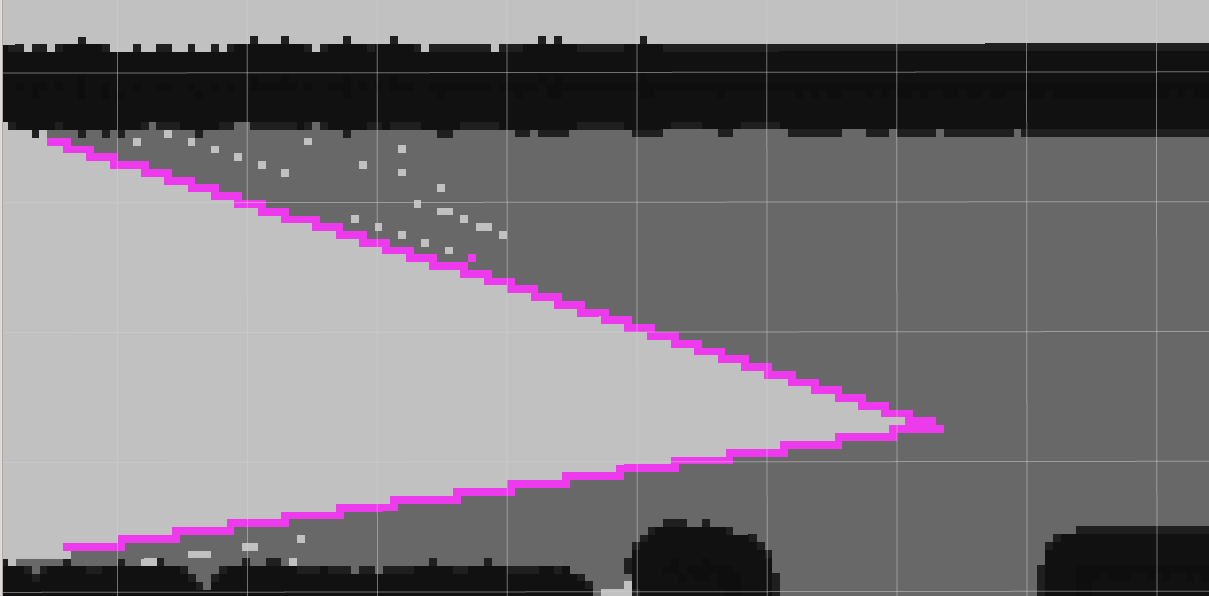}}
\\
\subfloat[Gazebo~\cite{gazebo} simulation of a practice environment with multiple pedestrians. A 3D lidar is mounted on top of the robot. Representations (c-e) are obtained from small variations of this environment. ]{\includegraphics[width=0.23\textwidth]{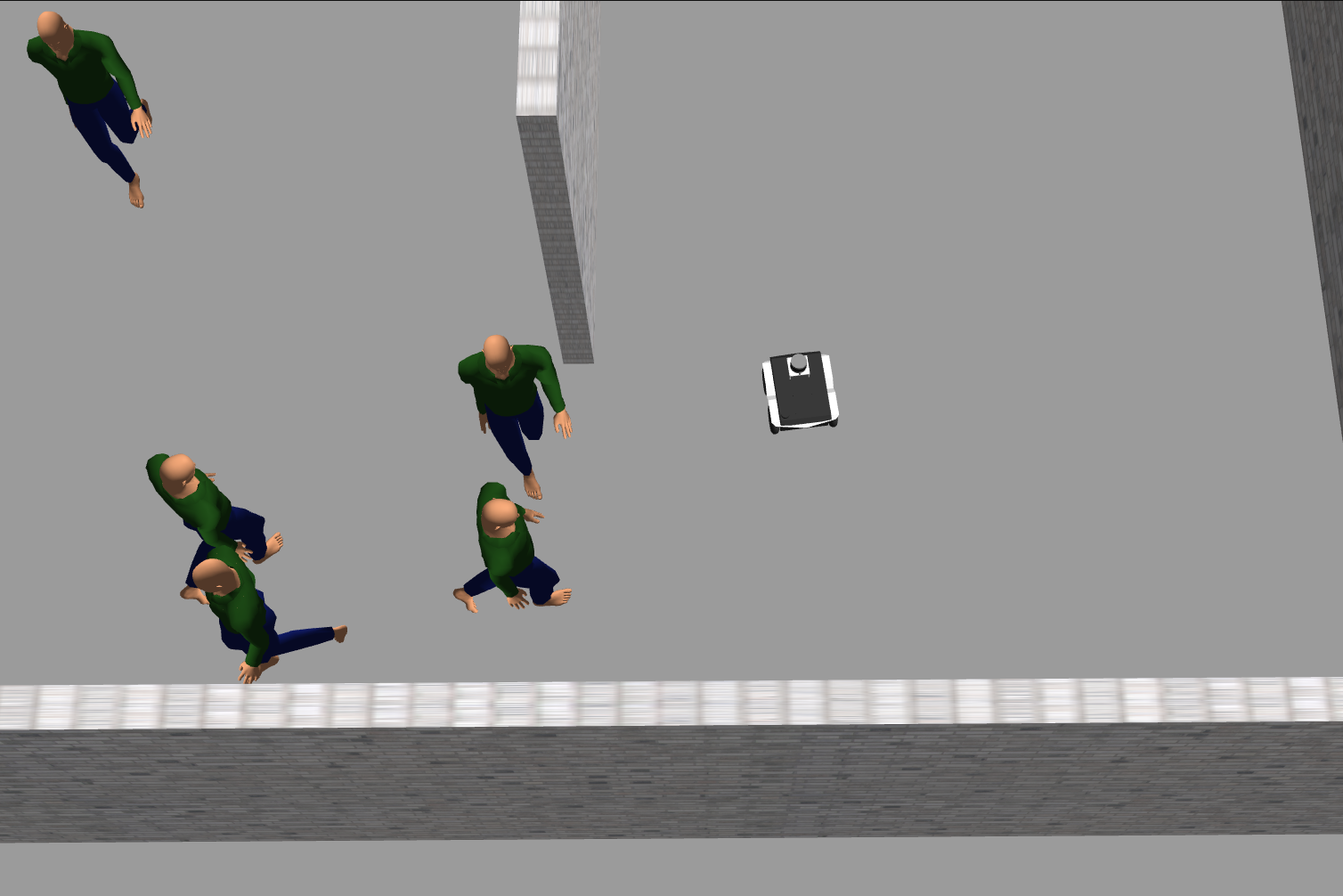}} \hfill
\subfloat[Frontier of type $mixed~simple$: $sd\_ratio = 20.7$]{\includegraphics[width=0.23\textwidth]{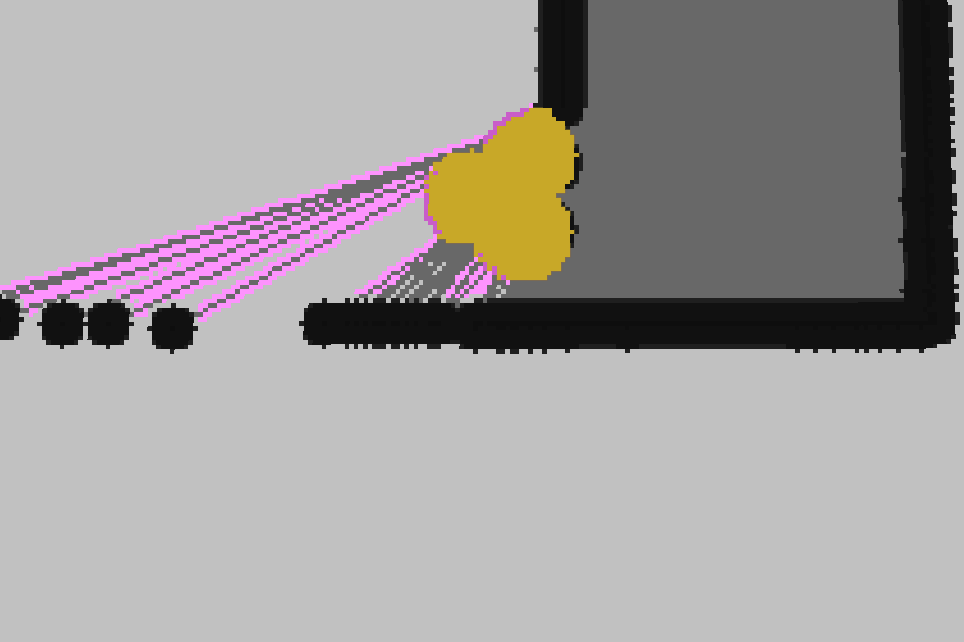}}
\\
\subfloat[Frontier of type $mixed$: $sd\_ratio= 0.70$]{\includegraphics[width=0.23\textwidth]{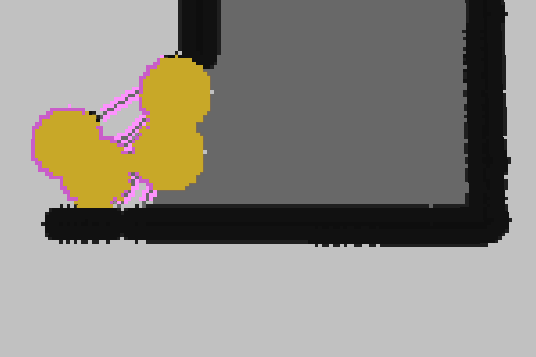}}
\hfill
\subfloat[Frontier of type $dynamic$: $sd\_ratio= 0.02$]{\includegraphics[width=0.23\textwidth]{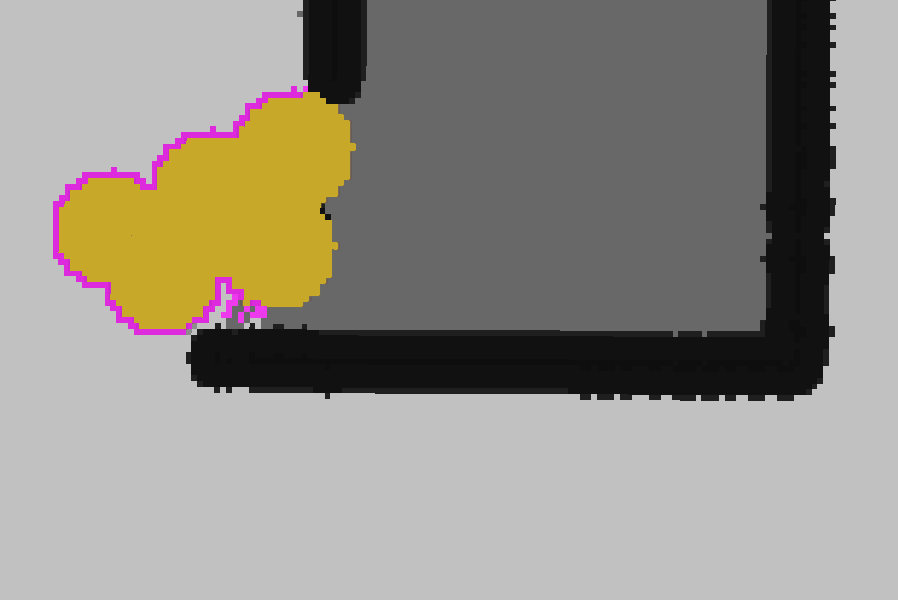}}
\caption{Illustration of different types of frontier defined in~\eqref{eq:dynamic-frontiers}. Type $simple$ of Subfig.~(a)  corresponds to the classic definition of frontier while Subfig.~(c-e) show the newly introduced types. Unknown cells are colored in light gray and free cells (ground) in the darker gray. Yellow blobs correspond to the moving actors shown in Subfig.~(b), inflated with a security radius. In black are the static obstacle cells and finally in bright pink the frontier cells.  For determining the type of each frontier a $thresh = 20$ is used.}
\label{fig:frontiers}
\end{figure}
    \item Decreases for frontiers of type $dynamic$ or $mixed$ with increasing elapsed time since first detection as it is likely that the dynamic obstacles have moved and freed the path since their last sight.
\end{itemize}
The cost function is formally defined in~\eqref{eq:cost}. The frontier with the lowest cost is selected (line~\ref{line:highest-pri}). Its travel point is computed (line~\ref{line:centroid}) as the frontier's centroid or the free cell closest to the robot and the originating frontier cell if the latter is a dynamic frontier cell. It is then set as next goal for the navigation module (line~\ref{line:send}). With the defined cost function, in most cases, $dynamic$ or $mixed$ frontiers would not be selected first as navigation goals if a $simple$ frontier was available. The robot would move closer to the selected goal and new closest frontiers would likely be found. The previously unchosen partly $dynamic$ frontiers could not be detected anymore for a long time because not closer than the closest $simple$ frontier.
To address this matter, we introduce a $memory$ that stores the location and characteristics (sizes, type and timestamp of first detection) of once detected $dynamic$, $mixed$ and $mixed~simple$ frontiers that were not selected as goals (line~\ref{line:add-to-memory}). Each new iteration, we update the $memory$ (line \ref{line:still-active}) by keeping the frontiers still present in the current map and not detected by the current frontier search (of line \ref{line:bfs-dyn}). If necessary, the frontiers characteristics are updated. These $memory$ frontiers are added to the list to be sorted and can therefore also be selected as the next goal (lines \ref{line:ascending}, \ref{line:highest-pri}), this could happen for example if enough time has passed or if the frontier's type has changed.
Finally, if no frontiers are detected nor still present in $memory$, the mapping is considered complete and the exploration terminates (line \ref{alg-fs-dyn:last-line}).
It is important to note that the approach is based on the assumption that dynamic objects will not maintain the same position over time but will eventually move, even by a small amount, leading to changed surroundings for the robot to explore.
\begin{algorithm}[t]
\caption{Goal Selection Procedure} \label{alg:frontier-search-dyn}
\begin{algorithmic} [1]
\ENSURE map, robot position in map $p$, memory
\STATE  $\text{candidates} \leftarrow $ frontierSearch($p$)\label{line:bfs-dyn}
\STATE memory $\leftarrow$ updateActiveMemory(memory, candidates) \label{line:still-active}
\STATE sorted\_frontiers $\leftarrow$ sort(\text{candidates} + \text{memory}) \label{line:ascending}
\STATE output\_frontier $\leftarrow$ sorted\_frontiers.pop() \label{line:highest-pri}
\IF{output\_frontier was found}
\STATE $ \text{goal} \leftarrow $ getTravelPoint(output\_frontier) \label{line:centroid}
\IF{is different goal}
\STATE sendGoal(goal) \label{line:send}
\ENDIF
\STATE addToMemory(sorted\_frontiers)\label{line:add-to-memory}
\ELSE
\STATE mapping is done \label{alg-fs-dyn:last-line}
\ENDIF
\end{algorithmic}
\end{algorithm}
\subsection{Cost Function}\label{subsec:cost}
Formally, based on the considerations of Section \ref{subsec:goal-selection-strategy}, the cost of a frontier $f$, with the current position of the robot $p$ and the travel point of the frontier $t$, is given by:
\begin{equation}\label{eq:cost}
    cost_f = \alpha \cdot \| p-t \| \ + \beta +  \gamma \cdot \frac{size\_d}{size\_t} \ + \delta \cdot (\zeta \cdot \Delta t^{\eta} + \theta \cdot oor)
\end{equation}
$\text{with coefficients} \ \alpha, \ \gamma, \ \eta > 0 $
$\text{and} \ \zeta, \ \theta < 0 $. 

The coefficient $ \beta $ depends on the frontier's type:
\begin{equation} \label{eq:cost-coeffs-type}
\beta = 
\begin{cases}
c_1  & \ \text{if type is } simple\\
c_2 > c_1 & \ \text{if type is } mixed \ simple \\
c_3 > c_2 & \ \text{if type is } mixed \\
c_4 > c_3 & \ \text{if type is } dynamic 
\end{cases}
\end{equation} 
The variable $size\_t$ corresponds to the total size of the frontier in number of cells. The boolean $\delta$ is false if the frontier is of type $simple$ or $mixed \ simple$, true otherwise. The quantity $\| p-t \|$ is the distance between the current robot position and the frontier's travel point.
The value of $\Delta t$ corresponds to the elapsed time since the frontier was first detected (only for types $mixed$ or $dynamic$). $oor$ is a boolean that indicates whether the travel point of the frontier is in the field of view of the robot or out of range.
The coefficients keep the formula dimensionless.

\section{Experiments}
The proposed cost based dynamic-aware strategy, CBD, is evaluated in two custom simulation environments featuring dynamic obstacles. Its performance is compared with our implementation of a closest frontier strategy, CF, that is not aware of dynamic objects \cite{frontier}. Our CF implementation features a BFS that scans only the known portion of the environment~\cite{wfd} as speed up. We also augment the original approach by recomputing the goal every time new information on the environment is available, as for our CBD strategy, to make the comparison fair. We expect other non dynamic-aware strategies to behave the same way. We use the Gazebo ROS simulation tool~\cite{gazebo} to simulate a jackal robot~\cite{jackal} equipped with a VLP-16 lidar mounted on top. The robot’s ground truth position and the position of moving objects around the robot are provided to our exploration module by Gazebo mimicking localization and dynamic object detection modules. Our pipeline includes the ROS navigation stack~\cite{navigation-stack}. \\
%
We conducted $24$ experiments in each of the two simulated environments and for each strategy. The strategies are compared on the obtained total path length, total duration, map divergence~\eqref{eq:divergence}, ineffective ratio, i.e. the total percentage of iterations that the robot spent without acquiring new knowledge of the environment and total loss~\eqref{eq:loss}.
In all runs, the minimum size a frontier needs to have to be detected is fixed at $30$ cells. The cost function coefficients~\eqref{eq:cost},~\eqref{eq:cost-coeffs-type} used for the experiments are specified in Table~\ref{table:best-params}.
\subsection{Loss} \label{subsec:loss}
The loss describes the overall quality of the exploration approach and we regard it as the performance measure to consider the most.
 Quantities of interest are the total path length and the total duration of the exploration. Also, any deviation from the ground truth map should be penalized. To measure the quality of a given map $map_a$ with respect to the ground truth map $map_{gt}$ we use the $map\_divergence$ which is defined as:~
\begin{equation}\label{eq:divergence}
    div(map_a, map_{gt}) = \frac{| (map_a \cup map_{gt}) -  (map_a \cap map_{gt}) |}{ | map_{gt} |}
\end{equation}

With $| \cdot |$ representing the cardinality of the set. The smaller the divergence, the more accurate the map (and exploration). In our experiments, the ground truth map has been manually selected as the most complete and visually accurate map of the simulation environment during previous exploration runs such that all areas have been observed. We define the loss of an exploration run $a$ as a weighted sum of the following quantities:

\begin{equation}\label{eq:loss}
loss_a =  \frac{tot\_length}{l} + \frac{tot\_time}{t} + \lambda \cdot div(map_a, map_{gt})
\end{equation}

Where $l$ is an estimated lower bound for the total path length $tot\_length$ and $t$ an estimated lower bound for the total exploration time $tot\_time$. The estimation of the lower bounds can be obtained in an independent set of experiments in the same environment by retaining the best observed outcome. We will use $\lambda = 20$ in our experiments as we value more having a complete map than the path length and time needed to achieve so.


\begin{table*}
\ra{1.4}
\centering
\caption{Summary of hospital and office-like environment results averaged over $24$ runs}
\label{table:results}
\begin{tabular}{lccccccc}\toprule
\multicolumn{1}{l}{} & \multicolumn{3}{c}{Hospital-like environment} & \phantom{abc} & \multicolumn{3}{c}{Office-like environment} \\
\cmidrule{2-4} \cmidrule{6-8}
 & CBD  & CF & Ratio [-]& & CBD & CF & Ratio [-]  \\ 
\midrule
\multicolumn{1}{l|}{Duration [s]} & 331.2 $\pm$ 135.2  & 75.1 $\pm$ 48.2  & 4.41     && 482.7  $\pm$ 85.3 & 309.9  $\pm$ 144.5 & 1.59 \\ 
\multicolumn{1}{l|}{Length [m]} & 220.5 $\pm$ 43.6   & 53.3 $\pm$ 35.2  & 4.14     && 300.8 $\pm$ 52.1 & 193.4  $\pm$ 92.6 & 1.56\\
\multicolumn{1}{l|}{Ineffective Ratio [-]} & 0.229 $\pm$ 0.137 & 0.19 $\pm$ 0.089 & 1.15 && 0.236 $\pm$ 0.046 & 0.395 $\pm$ 0.057 & 0.60 \\ 
\multicolumn{1}{l|}{Map divergence [-]} & 0.081 $\pm$ 0.138 & 0.659 $\pm$ 0.121 & 0.12 && 0.064   $\pm$ 0.057 & 0.292 $\pm$ 0.241 & 0.22 \\ 
\hline
\multicolumn{1}{l|}{Loss [-]} & \textbf{3.83} $\pm$ 2.63   & 13.69 $\pm$ 2.09 & 0.28  && \textbf{4.40} $\pm$ 1.09     & 7.85 $\pm$ 4.58 & 0.56      \\
\bottomrule
\end{tabular}
\end{table*}
\begin{figure}[t]
\centering
\subfloat[Hospital-like environment populated by 9 pedestrians.]{\includegraphics[width=0.83\linewidth]{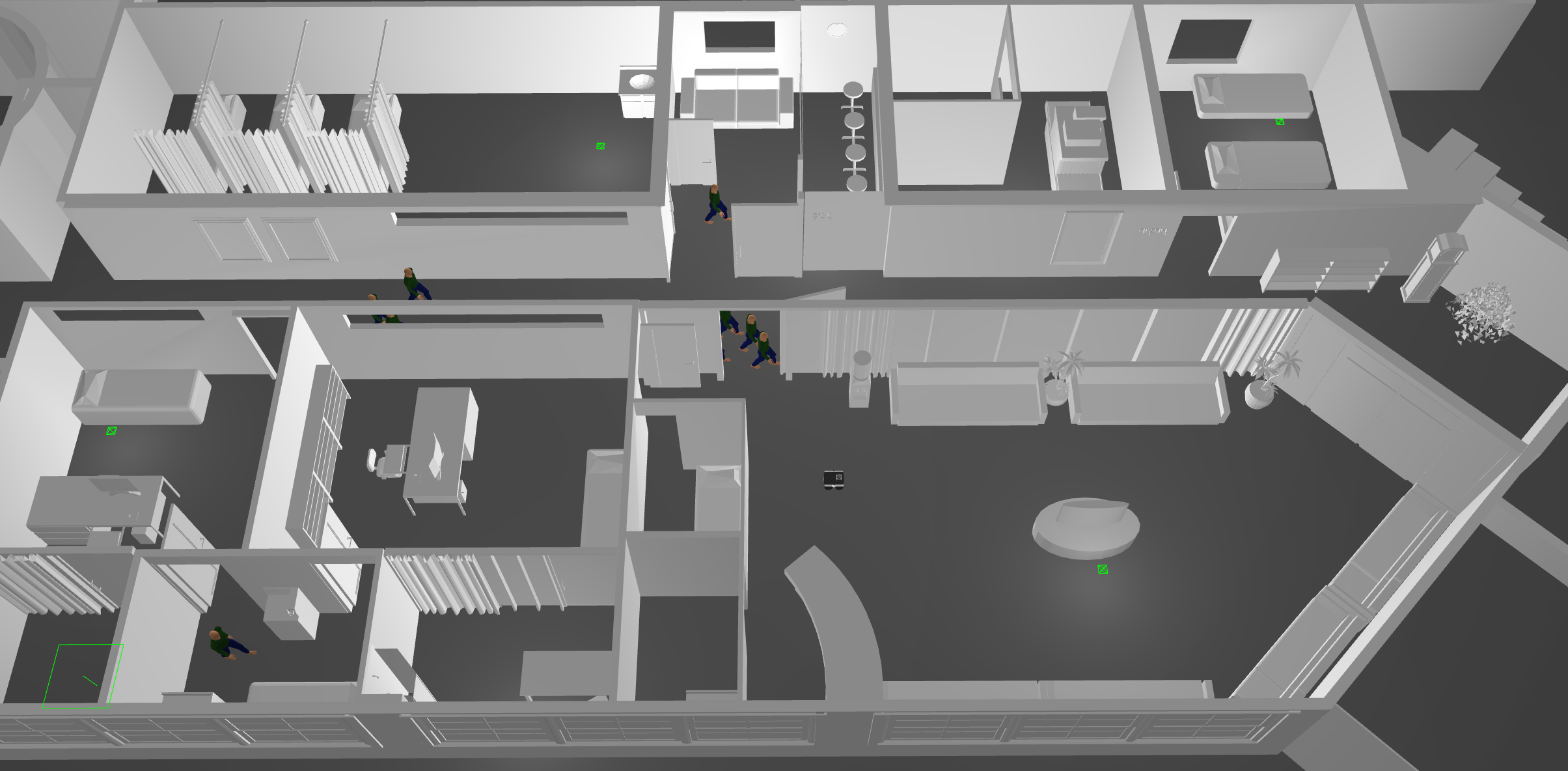}
} \\
\subfloat[Office-like environment populated by 13 actors. ]{\includegraphics[width=0.83\linewidth]{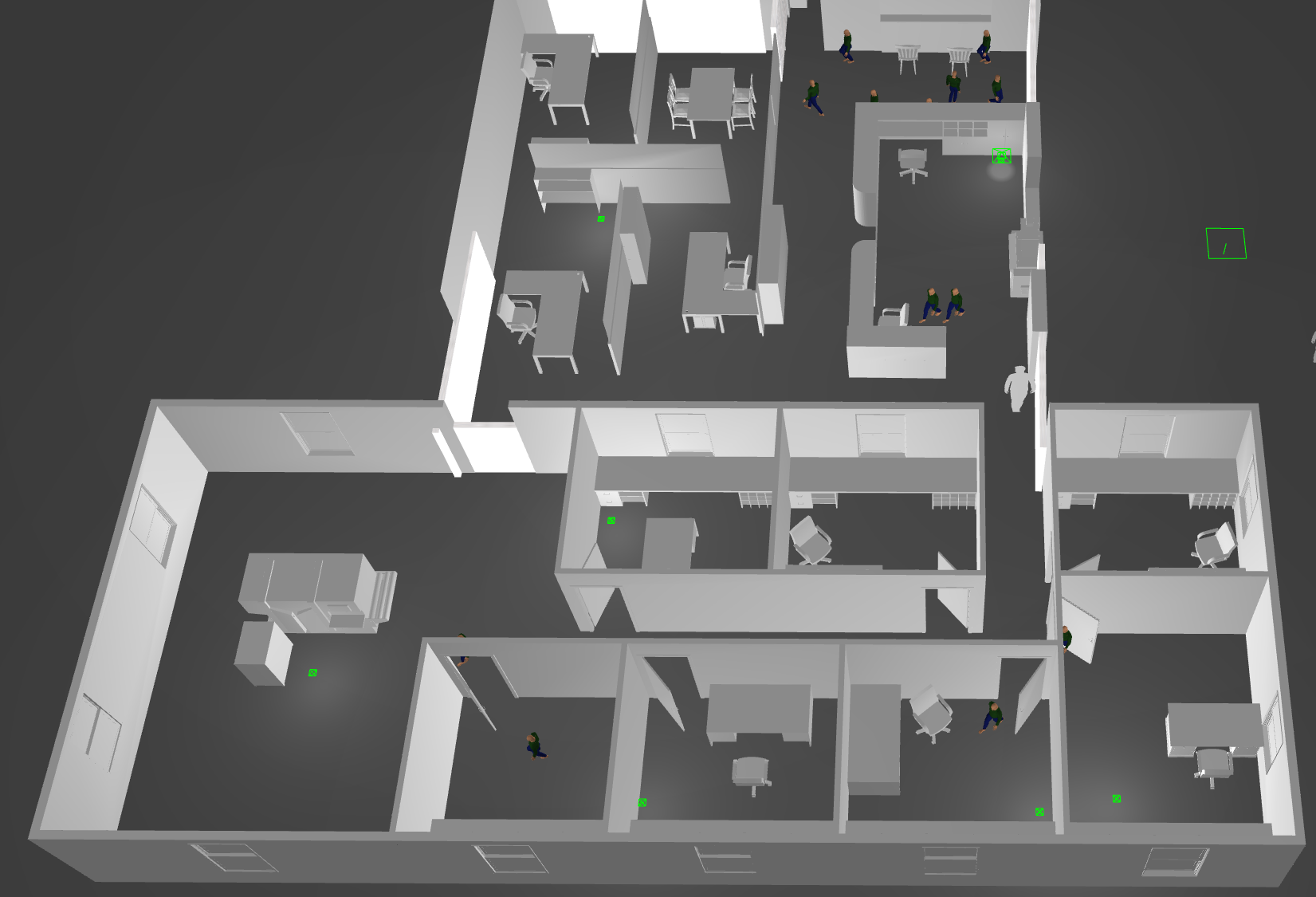}}
\caption{Partial view of the Gazebo simulated environments in which pedestrians move around following scripted trajectories. Note that green points simply represent light sources.}
\label{fig:environments}
\end{figure}
\subsection{Hospital-like Environment}
\begin{figure}[t]
\centering
\includegraphics[width=0.8\linewidth]{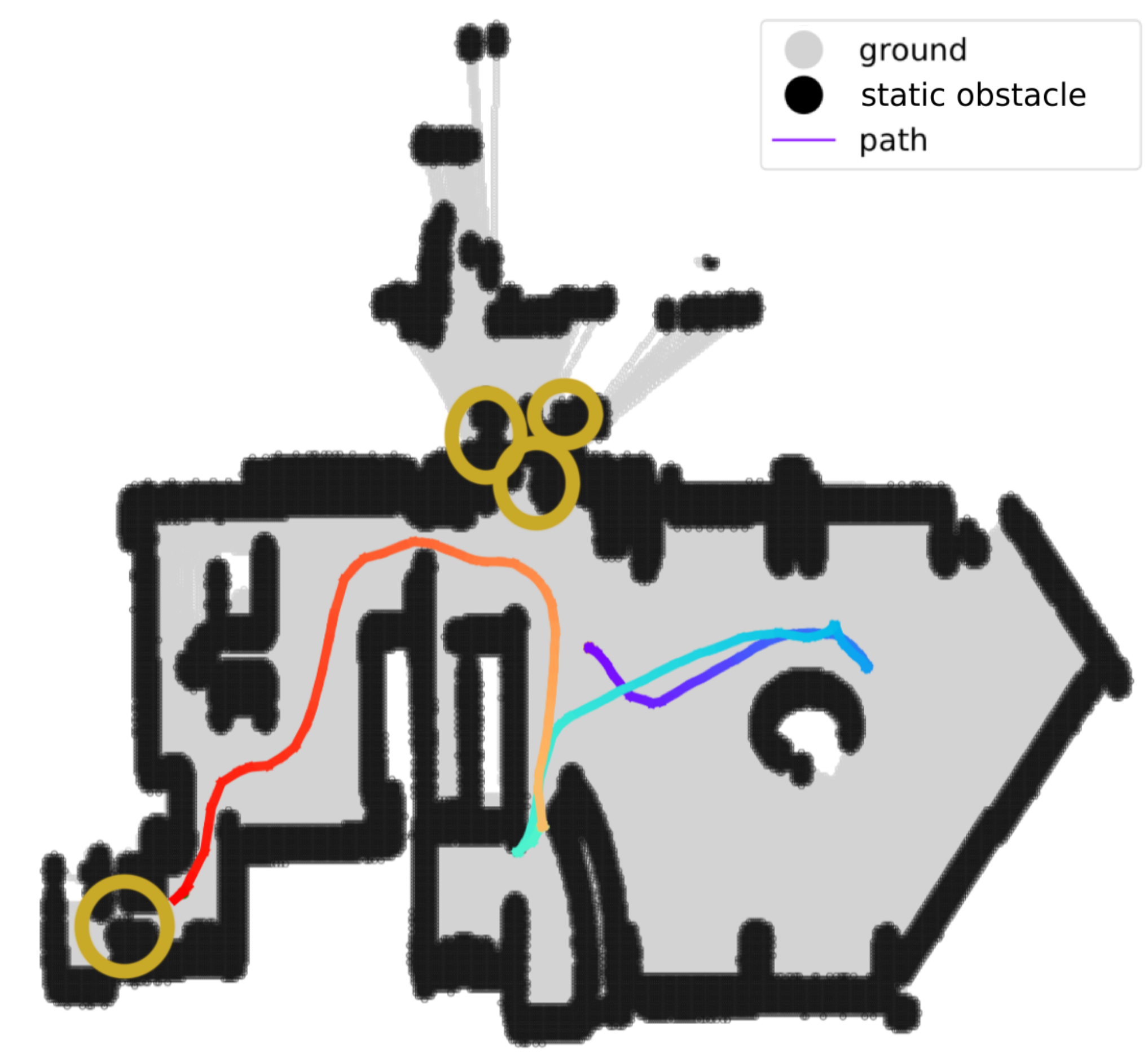}
\caption{Example run in the hospital-like simulation environment for the Closest Frontier (CF) exploration strategy. The path is colored with a rainbow colormap from the starting position in blue to the final one in red. Yellow circles have been added as an estimate of where the dynamic obstacles were when the robot observed that part of the environment.}
\label{fig:hosp-run-cf} 
\end{figure}

\begin{figure}[t]
\centering
\includegraphics[width=1.0\linewidth]{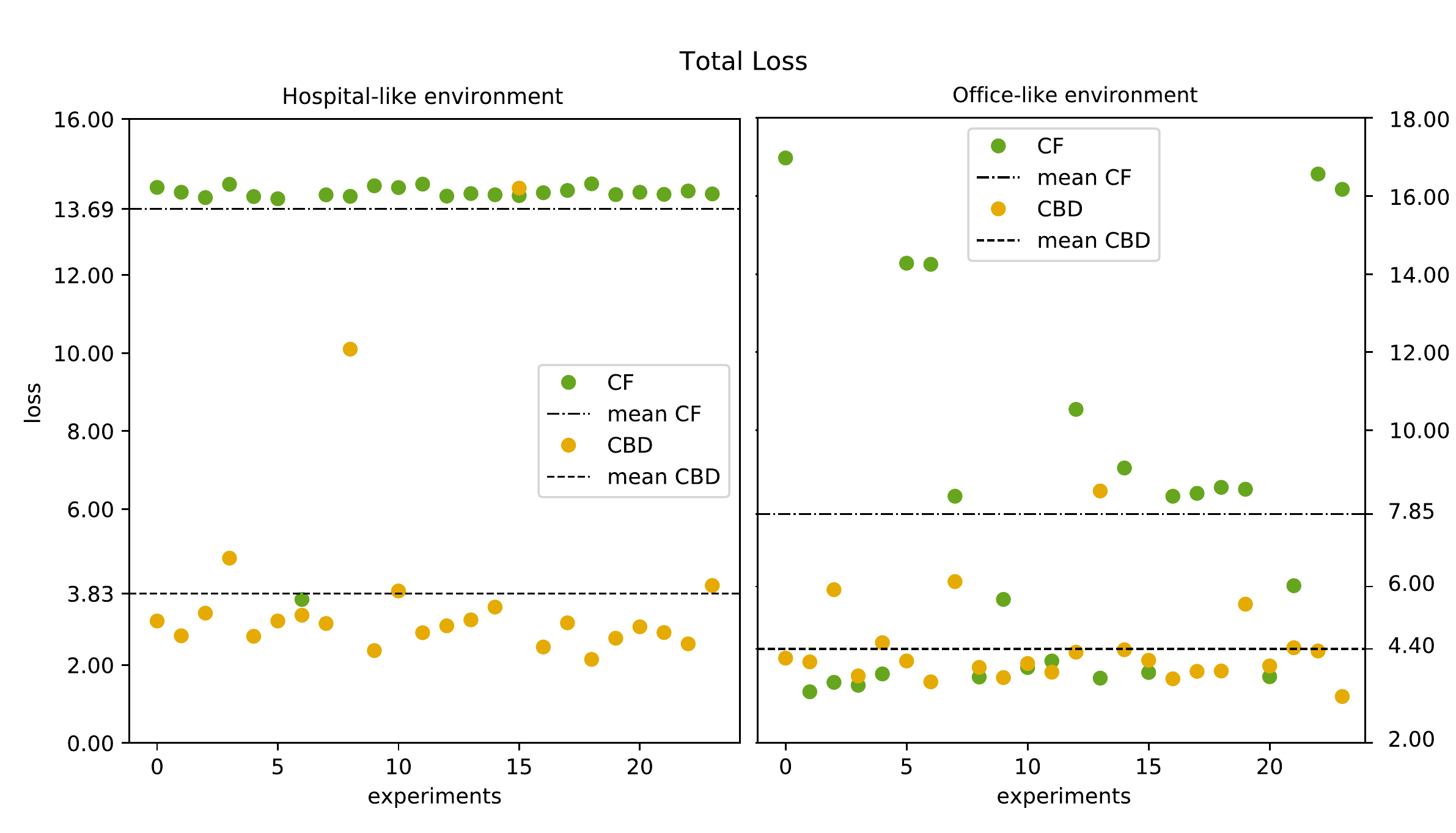}
\caption{Total $loss$ for each of the $24$ experiment runs in the two evaluated simulation environments, hospital-like and office-like. The $loss$ is calculated according to~\eqref{eq:loss}, with $ \lambda = 20$.} \label{fig:tot_loss}
\end{figure}

In the simulated hospital environment, shown in Fig. \ref{fig:environments}, people walk past the two entrances to the room in which the robot begins its exploration. It proves to be difficult for an exploration strategy not aware of dynamic obstacles (CF) to make the robot escape this room. Indeed, mapping only generic obstacles, the robot can exit the room if the door happens to be free during the time it is exploring that specific area. All $24$ experiments follow this trend, an example of which is shown in Fig. \ref{fig:hosp-run-cf}. Instead, our CBD strategy (cf. Fig.~\ref{fig:teaser}) is able to detect that the passage is blocked by dynamic obstacles and keeps them in memory while exploring the rest of the room. Eventually, the robot returns to the door's location to discover the passage as free. A summary of the complete results is available in Table \ref{table:results}. Note that the mean values of the total duration and exploration path length over all experiments are considerably lower for the CF strategy, due to the premature termination of exploration.
The approaches largely differ also in the final map divergence, caused by the incompleteness of the exploration using the CF approach.
Finally, the $loss$ confirms the superiority of the CBD strategy and is shown in Fig.~\ref{fig:tot_loss} for all $24$ runs.

\subsection{Office-like Environment}
The second environment for our evaluation is a simulation of an office featuring multiple small rooms and an open space with a big common room. A section of this environment is shown in Fig. \ref{fig:environments}. People walk around in the office covering inter and intra-room movement patterns. The $loss$ plot for all $24$ experiments for the two strategies is shown in Fig.~\ref{fig:tot_loss} and the summary of results is in Table \ref{table:results}.
The proposed CBD strategy achieves consistently a lower map divergence at the expense of a longer total path and duration when compared to CF strategy. The CBD method also utilizes the time spent for the mission more efficiently by spending less time without acquiring new knowledge as indicated by the ineffective ratio. In addition, the value of the $loss$ confirms the superiority of the proposed approach in this environment.
Note that experiments in the loss plot of Fig. \ref{fig:tot_loss} appear more scattered over the possible result space for CF strategy. This follows from the variety of its exploration's outcomes: some runs of CF strategy were successful in exploring the entire office, others terminated prematurely at different moments depending on the position of dynamic obstacles at the time of the robot's travelling close to critical locations (doors or entrances to corridors). The unreliability of CF strategy is also confirmed by noting the much higher standard deviation of all the results of Table \ref{table:results}.\\
Finally, it should be noted that in a system with no access to ground truth, the object detection module would likely produce false negatives and positives. In case the latter were persistent in location and time, with our CBD strategy the robot could be lead to explore the environment following a sub-optimal path but the exploration would nevertheless complete. With respect to false negatives instead, in a worst-case scenario, the proposed CBD approach would behave as a non-dynamic aware strategy.
\subsection{Optimization}
The cost function presented in~\eqref{eq:cost} features many coefficients that have to be tuned carefully.
The overall path length and total exploration time vary greatly depending on their values. The tuning problem can be formulated as an optimization problem. With $f = loss$  defined in~\eqref{eq:loss}, we want to solve: 
$x^\ast = \underset{x}{\mathrm{argmin}} \  f(x)$,  where $x$ represents a set of cost function coefficients from~\eqref{eq:cost}-\eqref{eq:cost-coeffs-type}, $~x^\ast$ the optimum set of coefficients for an (any) exploration task. The $loss$ as a function of the coefficients is treated as a blackbox function: the exact form is not known but it is possible to punctually evaluate it for one set of parameters $x$ by running a simulation and obtaining all the needed quantities. Through a sequential model-based optimization approach~\cite{optimization} we optimize the coefficients for the hospital-like environment and we show improvement in the obtained $loss$ (cf. Table \ref{table:best-params}), suggesting that it is possible to automatically tune the parameters for a given task. However, the optimization should be performed on data collected in diverse environments to avoid overfitting. We expect our hand-engineered parameter values to generalize better to other environments than the ones obtained using such an automated tuning approach on the single hospital-like environment.
\begin{table}
\caption{Cost function coefficients of~\eqref{eq:cost} and \eqref{eq:cost-coeffs-type} with the obtained loss in the hospital-like environment averaged over $24$ runs.}
\label{table:best-params}
\centering
\ra{1.3}
\begin{tabular}{@{}l|cc@{}} 
         & Proposed & Tuned (Opt)  \\
\midrule
$\alpha$  & 3.0      & 4.411  \\
$\gamma$ & 0.2      & 8.761  \\
$\zeta$  & -0.8     & -8.172 \\
$\eta$   & 1.5      & 4.332  \\
$\theta$ & -0.2     & -2.067 \\
$c_1$    & 5.0      & 0.129  \\
$c_2$    & 7.0      & 28.352 \\
$c_3$    & 60.0     & 43.275 \\
$c_4$   & 60.0     & 45.014 \\
$thresh$ & 20 & 20 (fixed) \\
\hline
Loss    & 3.833    & \textbf{3.2604} \\
\bottomrule
\end{tabular}
\end{table}

\section{Conclusion}
In this work we have proposed a new dynamic-aware exploration strategy that builds upon the well researched and widely applied concept of frontier-based exploration.
By explicitly taking into account dynamic obstacles at exploration time and including them in the strategy, we are able to achieve consistent and reliable exploration in unknown populated environments.
By designing an appropriate cost function, the exploration of frontiers that include dynamic obstacles can be postponed to a more suitable moment and, instead of waiting for such obstacles to free a blocked path, the robot may continue its exploration elsewhere.
The validity of the proposed approach has been shown and evaluated in two complex simulation environments that include challenging situations in which our baseline solution was not able to achieve satisfactory results. Finally, an automated tuning pipeline for the cost function coefficients has been suggested and tested. In future work we aim to build on top of the proposed cost function by integrating the behaviour of different categories of moving obstacles.
\addtolength{\textheight}{-0cm}
\bibliographystyle{IEEEtran}
\bibliography{references.bib}

\end{document}